\pdfoutput=1
\documentclass[preprint]{elsarticle}

\usepackage{amssymb}
\usepackage{amsmath}
\usepackage{array}
\usepackage{enumerate}
\usepackage{graphicx}
\usepackage{float}
\usepackage{etoolbox}
\usepackage{amsfonts,amssymb}
\usepackage{mathptmx}
\usepackage{txfonts}
\usepackage{mathrsfs}
\usepackage{subfigure}
\usepackage{graphicx}
\usepackage{booktabs}
\usepackage{algorithm}  
\usepackage{algorithmicx}
\usepackage{algpseudocode}
\usepackage{threeparttable}
\usepackage{mathtools}
\usepackage{color}
\usepackage{pifont}
\usepackage{hyperref}
\hypersetup{hidelinks,
	colorlinks=true,
	citecolor=blue,
	linkcolor=blue}

\journal{none}

\begin{document}
\begin{frontmatter}



\title{Anchor Free remote sensing detector based on solving discrete polar coordinate equation}


\author[1]{Linfeng Shi \corref{cor1}}
\ead{20211249579@nuist.edu.cn}

\author[1]{Yan Li}
\ead{002200@nuist.edu.cn}

\author[1]{Xi Zhu}
\ead{20211249603@nuist.edu.cn}

\address[1]{Automated institute, Nanjing University of Information
	Technology, No. 219, Ningliu Road, Nanjing, 210000, Jiang Su, China}
\cortext[cor1]{Corresponding author}
\begin{abstract}
As the rapid development of depth learning, object detection in aviatic remote sensing images has become increasingly popular in recent years. Most of the current Anchor Free detectors based on key point detection sampling directly regression and classification features, with the design of object loss function based on the horizontal bounding box. It is more challenging for complex and diverse aviatic remote sensing object. In this paper, we propose an Anchor Free aviatic remote sensing object detector (BWP-Det) to detect rotating and multi-scale object. Specifically, we design a interactive double-branch(IDB) up-sampling network, in which one branch gradually up-sampling is used for the prediction of Heatmap, and the other branch is used for the regression of boundary box parameters. We improve a weighted multi-scale convolution (WmConv) in order to highlight the difference between foreground and background. We extracted Pixel level attention features from the middle layer to guide the two branches to pay attention to effective object information in the sampling process. Finally, referring to the calculation idea of horizontal IoU, we design a rotating IoU based on the split polar coordinate plane, namely JIoU, which is expressed as the intersection ratio following discretization of the inner ellipse of the rotating bounding box, to solve the correlation between angle and side length in the regression process of the rotating bounding box. Ultimately, BWP-Det, our experiments on DOTA, UCAS-AOD and NWPU VHR-10 datasets show, achieves advanced performance with simpler models and fewer regression parameters.
\end{abstract}


\end{frontmatter}


\section{Introduction}\label{sec1}
The emergence of deep convolutional neural networks has promoted the rapid development of object detection. Many object detectors based on convolutional neural networks have achieved good results in natural scenes. Different from traditional natural images, objects in aviatic remote sensing images\cite{xia2018dota,li2019feature,cheng2016learning} are shot from the aviatic perspective. The adoption of horizontal detectors would bring a series of problems\cite{yang2022scrdet++}: 1) A large field of vision in aviatic remote sensing images, containing a variety of background information more often than not, would interfere with the extraction of object features. 2) As many small objects in aviatic remote sensing images, the Pooling layer will further reduce the amount of information, leading to the loss of small object information in deep features. 3) As the shooting angle looking down from the air, the different observation mode of the object, and the uncertain direction of the object, it is hardly robust to the direction of the remote sensing object on the conventional horizontal detector.

In view of the problems of the above horizontal detectors, some researches have designed the detection algorithm of the directional object. The two-stage algorithm locates the rotating rectangular box by adding the rotation angle parameter\cite{jiang2017r2cnn,ma2018arbitrary}, or it is improved the loss function to regress the four corners to locate the polygonal boundary box\cite{xu2020gliding}. It is the same for the one-stage remote sensing object detection algorithm. TextBoxes++ represents a single-stage model first applied to text detection\cite{liao2018textboxes++}. On the remote sensing object detection task, R3Det designed a feature fine tuning module based on RetinaNet algorithm for remote sensing object\cite{yang2021r3det}.

\begin{figure}[h]%
	\centering
	\subfigure[]{
		\begin{minipage}[t]{0.58\linewidth}
			\centering
			\includegraphics[width=3.1in]{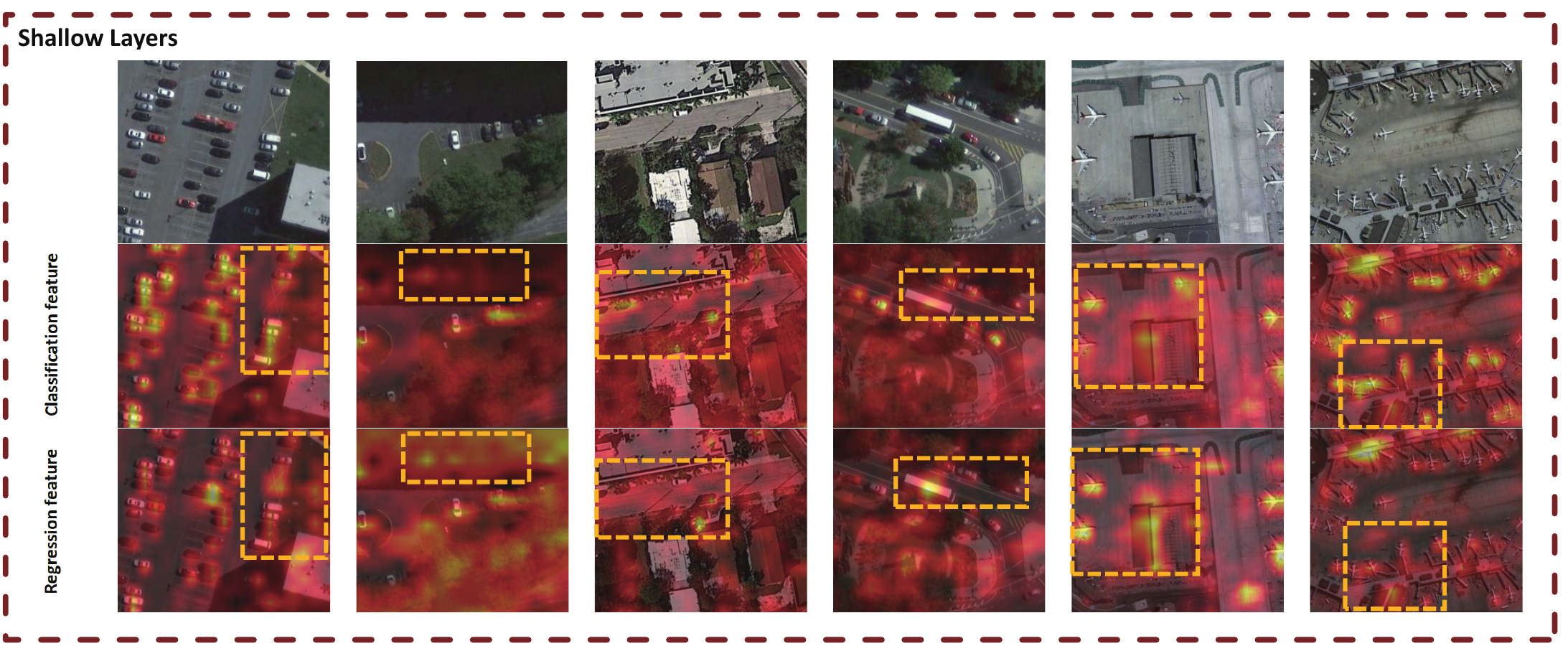}
			%\caption{fig2}
		\end{minipage}%
	}%
	\subfigure[]{
		\begin{minipage}[t]{0.5\linewidth}
			\centering
			\includegraphics[width=1.6in]{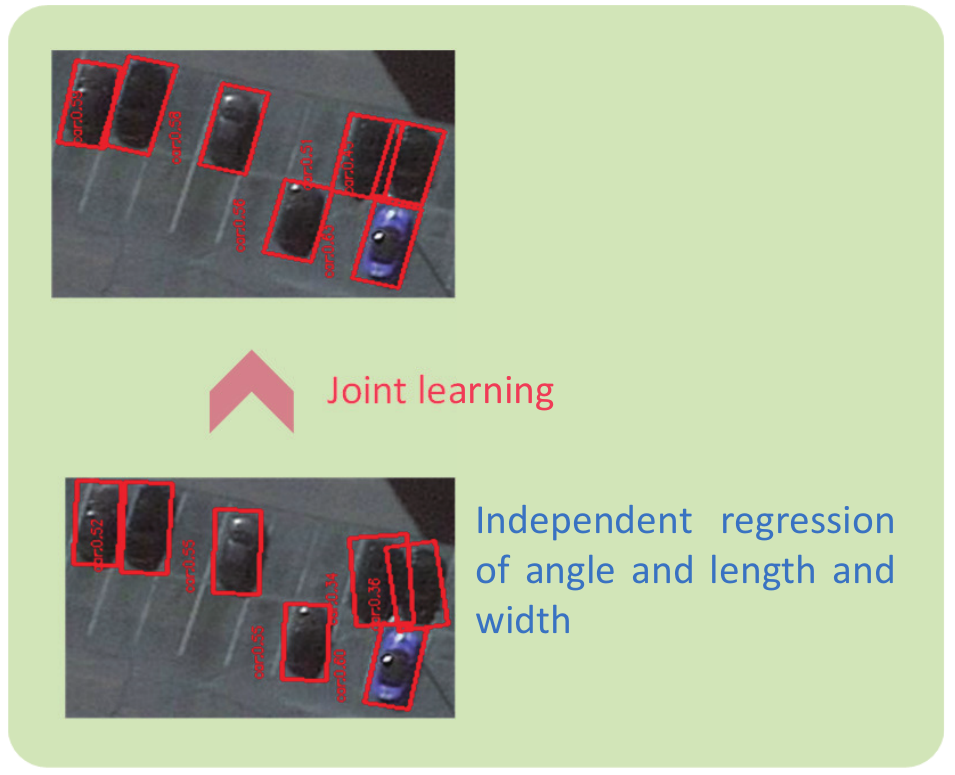}
			%\caption{fig2}
		\end{minipage}%
	}%
	\caption{(a) Sensitivity map of regression and Heatmap classification task characteristics in Anchor Free model. (b) The regression is unstable due to the loss of each part of the parameters calculated separately.}\label{fig1}
\end{figure}

Notwithstanding satisfactory performance achieved from the above methods, it is need to design the complex Anchor to improve the detection accuracy of the object, the detection results are very sensitive to the design of the Anchor. In addition, the existing Anchor Free model is based on baseline\cite{he2016deep,newell2016stacked}, which transforms the angle into the prediction of key points\cite{wei2020oriented} or vectors\cite{yi2021oriented} by designing different directional boundary box representation methods. These methods rarely consider the independent and cooperative relationship of classification and regression in the process of model up-sampling,  We find that this will cause feature coupling and weaken the semantic information of the two subtasks. Figure \ref{fig1}a shows the feature sensitivity map of the Anchor Free model classification and regression tasks. It can be found that different subtasks focus on different information. In the classification task, the Heatmap learns more detailed object location and classified local information. While in the regression task, it pays more attention to the global information related to the object. At the same time, in the process of up-sampling, the complex background of remote sensing image will also bring more noise information of different scales.

Due to the particularity of the shooting angle of the remote sensing image, the object has the problem of direction angle. As shown in Figure \ref{fig1}b, The loss function designed in the past is to calculate the loss of each part of the parameters of the boundary box separately, with each parameter not related to each other, and the inconsistent measurement units of angle and length and width will lead to over-fitting of local parameters. It leads to unstable regression of the model, thus reducing the convergence performance of the network. To sum up the above, we propose a way to reasonably allocate and fuse deep and shallow feature information, design a new enhancement channel and spatial feature extraction module to highlight the difference between foreground and background, and jointly study the loss function of angle and length and width information.

\begin{figure}[h]%
	\centering
	\includegraphics[width=0.8\textwidth]{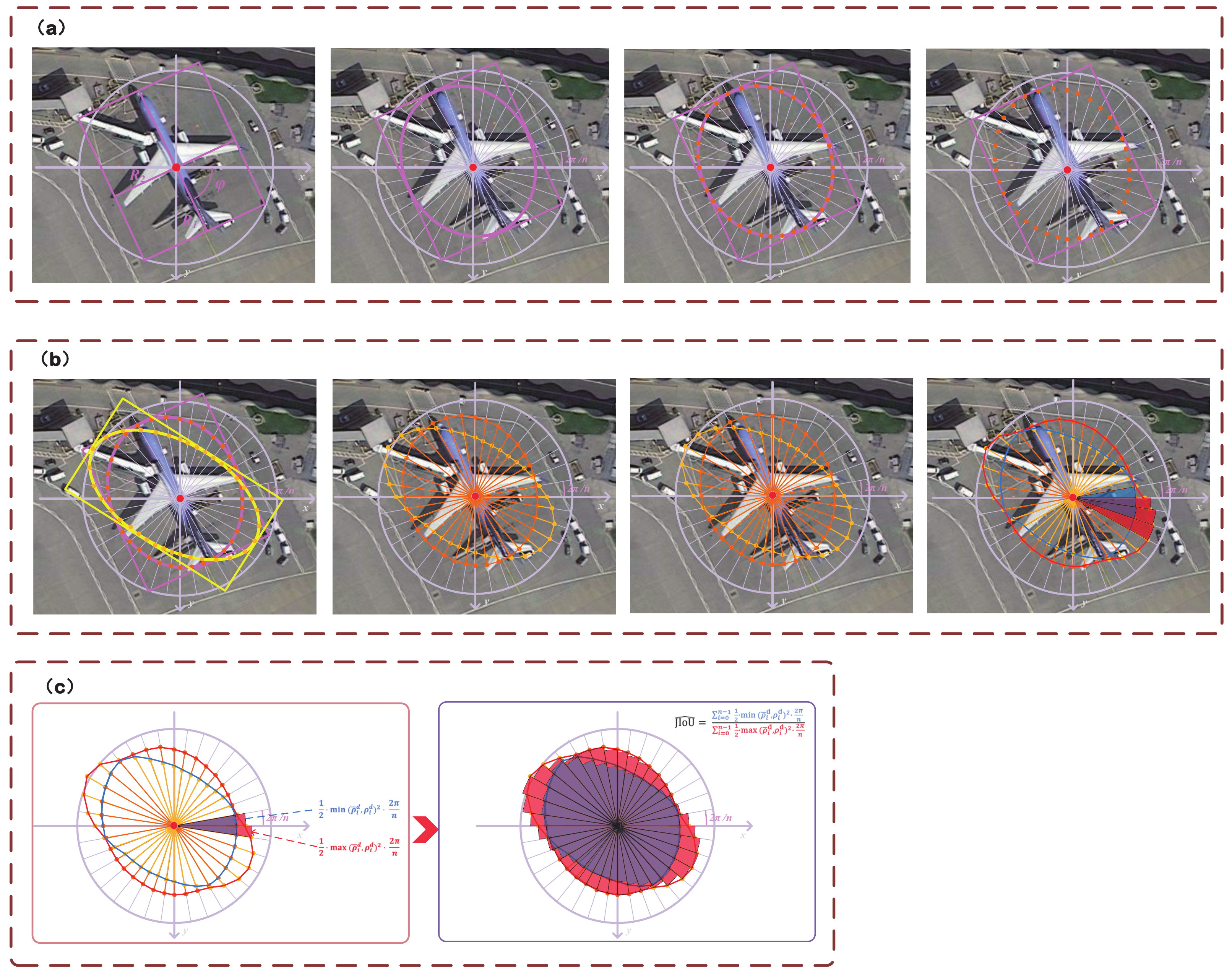}
	\caption{(a) shows the conversion method of directional bounding boxes to discrete ellipses, with (b) showing the approximate IoU calculation method of two directional bounding boxes. In the figure, pink refers to the ground truth, with yellow representing the prediction box. The prediction box along with ground truth are expressed using the elliptical polar coordinate formula. It is obtained on the intersection and merger specific area of two discrete ellipses following discretization. Blue represents the union of two discrete ellipses, with red representing the intersection of two ellipses. (c) Visualization of JIoU calculation process.}\label{fig2}
\end{figure}

We propose a novel Anchor Free remote sensing detector based on solving discrete polar coordinate equation, which is mainly used to solve the following problems:
\begin{enumerate}[1.]
	\item A interactive double-branch up-sampling network is proposed, which reasonably distributes and integration deep and shallow features. In the up-sampling process, middle layer features are fused and a new branch is introduced to regress the boundary box parameters.
	\item An approximate IoU loss function (JIoU) based on polar coordinate segmentation is proposed to jointly learn the direction, length and width parameters of the bounding box for the problem of non correlation between angle and length and width in the directional bounding box regression process, and design experiments to verify the effectiveness of JIoU. 
	\item To solve the problem of noise in remote sensing images, we introduce weight coefficients into multi-scale convolution to generate pixel-level attention features and perform visual analysis.
	\item It is validated the performance of the model on multiple datasets, as shown in Table \ref{tab3}. On the DOTA\cite{xia2018dota} dataset. With ResNext50 as the BackBone, the mAP reaches 71.42$\%$. Table \ref{tab2} and Table \ref{tab1} show the results on the UCAS-AOD\cite{li2019feature} and NWPU VHR-10\cite{cheng2016learning} datasets, with mAP 97.88$\%$ and 92.07$\%$ respectively. Table \ref{tab4} shows the results of the ablation experiment on UCAS-AOD. When ResNext50 remains BackBone, the FPS represents 13.83. After ResNet18 is replaced, the FPS is 46.29 and the mAP reaches 96.37$\%$.
\end{enumerate}
\section{Related work}\label{sec2}
Various object detectors have been described in many literatures, which can be divided into two categories as per their detection structures, namely, horizontal detectors and directional detectors. This section will explain the contributions of previous scholars in horizontal detection and directional detection.
\subsection{Horizontal object detection}\label{subsec1}
Various level detectors have been designed at present, which can be divided into Anchor Free model and Anchor Based model as per different Anchor mechanisms. 

The two-stage model RCNN generates Possible Regions by detecting ROI (Region Of Interest), sending each Possible Region into the network to learn features. These features are passed to the classifier to identify object categories\cite{girshick2014rich}. Fast RCNN designed ROI pooling layer to improve RCNN through features learned from the whole picture through the network\cite{girshick2015fast}. A RPN (Region Proposal Networks) is designed in Faster RCNN to generate Region Proposal, which speeds up the processing of two-phase models\cite{7485869}. The single junction model YOLO directly predicts the coordinates of the bounding box, the confidence level of the object and the probability of the object condition category through the network\cite{redmon2016you}. YOLO9000 further improves the accuracy and speed of the network by introducing the Anchor mechanism and other strategies\cite{redmon2017yolo9000}. YOLOV3 improves the detection accuracy of small object by integrating features of different scales\cite{redmon2018yolov3}. 

In the above-mentioned literature, it is necessary to design complex Anchor. In contrast, the Anchor Free method is more concise, in which FCOS performs classification and regression by pixel prediction\cite{tian2019fcos}. CornerNet generates a Heatmap by detecting the corners of the object's bounding box\cite{law2018cornernet}. It is obtained on the location and category information of the object. CenterNet locates and classifies the object by detecting the object center point and offset, and regresses the boundary box parameters to follow\cite{zhou2019objects}. However, it is the complex background of remote sensing images, the diversity of object directions and large vertical and horizontal ratio that empower these horizontal object detectors to face many problems.
\subsection{Directional object detection}\label{subsec2}
Some scholars have proposed detection methods for remote sensing object in view of the problems faced by the horizontal object detectors. 

In the Anchor Based method, R2CNN uses the horizontal Anchor to extract RoI in the first stage, generating rotation candidate areas in the RPN stage and uses multi-scale pooling operation to enhance the generalization ability of detection and recognition features. In the second stage, it realizes the regression of the rotation box based on the horizontal candidate areas\cite{jiang2017r2cnn}. RRPN defines a large number of Anchor with direction angles for the regression of rotating object, proposing the pooling operation of rotation features for normalization in the reasoning phase\cite{ma2018arbitrary}. RoI Transformer inserts a lightweight module between RPN and RCNN. It converts the horizontal region generated by RPN into a rotating region and reduces the computational complexity of rotating object detection\cite{ding2019learning}. SCRNet, for the distribution of dense object, achieves feature enhancement of small object and dense object by training pixel attention map and increasing channel attention mechanism\cite{yang2019scrdet}. SCRDet++ guides the feature extraction process of detection recognition with the help of intermediate features of semantic segmenting networks, indirectly using the attention mechanism to enhance the features of dense object, so as to improve the degree of boundary discrimination of dense object\cite{yang2022scrdet++}. R3Det ensures that the features of the object are aligned with the center of the object by adding a feature map fine-tuning module for more accurate positioning of dense object\cite{yang2021r3det}. SCRNet designed the IOU Smooth loss function to directly weaken the boundary samples during training for the improvement of the loss function\cite{yang2019scrdet}. 

It is the complex design of Anchor that empower some scholars to propose Anchor Free's remote sensing object detection method. O2-DNet detects objects by predicting a pair of middle lines inside each object\cite{wei2020oriented}. P-RSDet introduced polar coordinate system into remote sensing object detection for the first time, proposing a polar ring area loss function to constrain the geometric relationship between polar radius and polar angle\cite{zhou2020arbitrary}. So did PolarDet to represent the directional object in polar coordinates. He designed a sub-pixel center semantic structure to further improve the classification accuracy\cite{zhao2021polardet}. BBAVectors detected the central key point of the object, and predicted the perception vectors to regress the directional bounding box to follow\cite{yi2021oriented}.
\section{Proposed method}\label{sec3}
Initially, we will introduce the Anchor Free directional detecting model proposed in this paper, subsequently, explain the weighted multi-scale convolution improved in this paper, and ultimately, elaborate the representation of boundary box parameters and the loss function of the model. BWP-Det follows the point detection structure to general key, using ResNext50 as the feature extraction network. The classification and regression branch up-sample the fused features separately to obtain different results, and generates pixel level attention by the improved weighted multi-scale convolution to guide the interactive double-branch up-sampling process.
\begin{figure}[h]%
	\centering
	\includegraphics[width=1\textwidth]{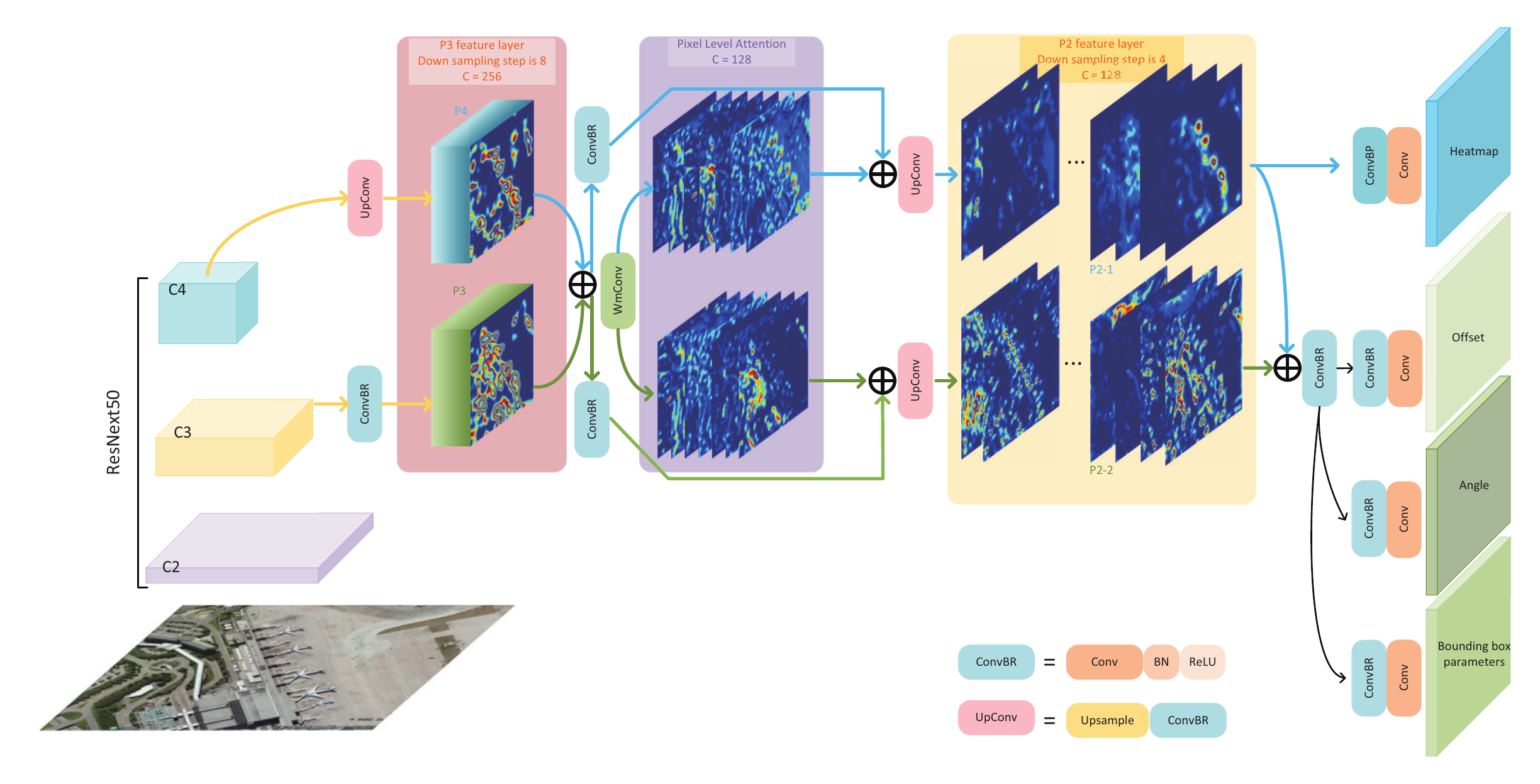}
	\caption{The overall network structure of BWP-Det uses two up-sampling branches for classification and regression, adds angle prediction branches to the detection head, and generates pixel level attention features based on weighted multi-scale convolution.}\label{fig3}
\end{figure}
\subsection{More Reasonable Up-sampling Method}\label{subsec3}
We did not choose to sample on the deep feature $\mathcal{C}$5 due to the use of pooled layers. Small objects in deep features lose most of their feature information, and larger sampling steps tend to directly skip smaller objects, leading to the loss of small objects.

In general, Anchor Free model directly samples deep features, and uses different detectors for regression and classification tasks to follow. However, this is not an effective feature allocation method.   The difference between classification and regression features can be seen from the display of classification and regression features in Figure $\ref{fig1}$a. The regression task pays more attention to the overall information of the target, but this may be unnecessary noise for the classification task. Similarly, the classification task pays attention to local information, which will weaken the characteristics of the regression task, and directly locating the regression features through hard labels will also cause the problem of incorrect features.

In addition, the sampling range of depth feature is larger, which will lead to more fuzzy positioning of small objects densely arranged in the Heatmap. The features with larger scales can make up for this defect. However, the shallow features with larger scales have insufficient semantic information. Not only is it hardly conducive to object classification, but leads to insufficient receptive field for larger objects. As such, we propose an interactive upsampling method 

Backed by the above analysis, the specific implementation method of interactive double-branch up-sampling is as follows. Figure \ref{fig4}a shows the overall network structure of this paper. $\mathcal{P}$4 is obtained by up-sampling through bilinear interpolation and joint convolution as starting from the sub deep layer feature  $\mathcal{C}$4.  $\mathcal{C}$3 is fused with $\mathcal{P}$4 after using convolution to change the number of channels. $\mathcal{PA}$ is extracted from the fused features through improved weighted multi-scale convolution. $\mathcal{PA}$ contains more object information and less background noise information, guiding the learning of the model. In the classification sub task, the fused features are added to $\mathcal{PA}$-1 after convolution decoupling to obtain the location and category information of object features, with $\mathcal{P}$2-1 obtained from the results for generation and extraction to the key Heatmap. In the regression branch, the attention feature $\mathcal{PA}$-2 is added to the fusion feature after decoupling convolution, wit the final regression subtask feature $\mathcal{P}$2-2 is obtained by fusing the result of up-sampling with $\mathcal{P}$2-1,. These features focus on more precise object edge information, which is more conducive to the regression task of the model. Finally, different detection heads and corresponding loss functions are used for training. As mentioned above, the network model designed in this paper has a very simple structure and can conduct end-to-end training.
\subsection{Weighted multi-scale convolution }\label{subsec4}
Due to the large amount of background noise in remote sensing images, the process of up-sampling will bring deep and shallow noise information, which will affect the positioning and classification results of key points. The parameter regression of object boundary boxes needs more boundary location information, with the background noise not conducive to the prediction of these location information. As such, this paper proposes a weighted multi-scale convolution to reduce the impact of noise information.

\begin{figure}[h]%
	\centering
	\subfigure[]{
		\begin{minipage}[t]{0.5\linewidth}
			\centering
			\includegraphics[width=3.2in]{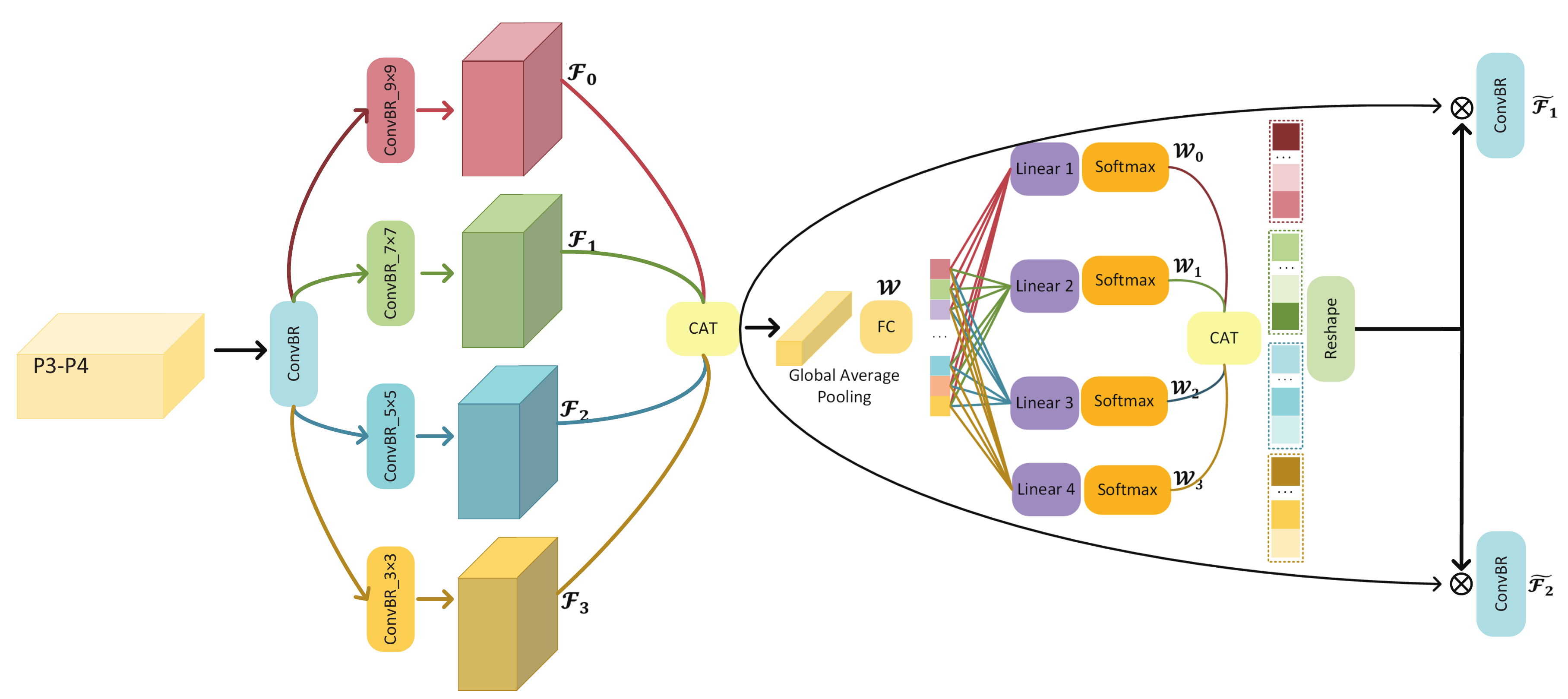}
			%\caption{fig2}
		\end{minipage}%
	}%
	\subfigure[]{
		\begin{minipage}[t]{0.7\linewidth}
			\centering
			\includegraphics[width=1.2in]{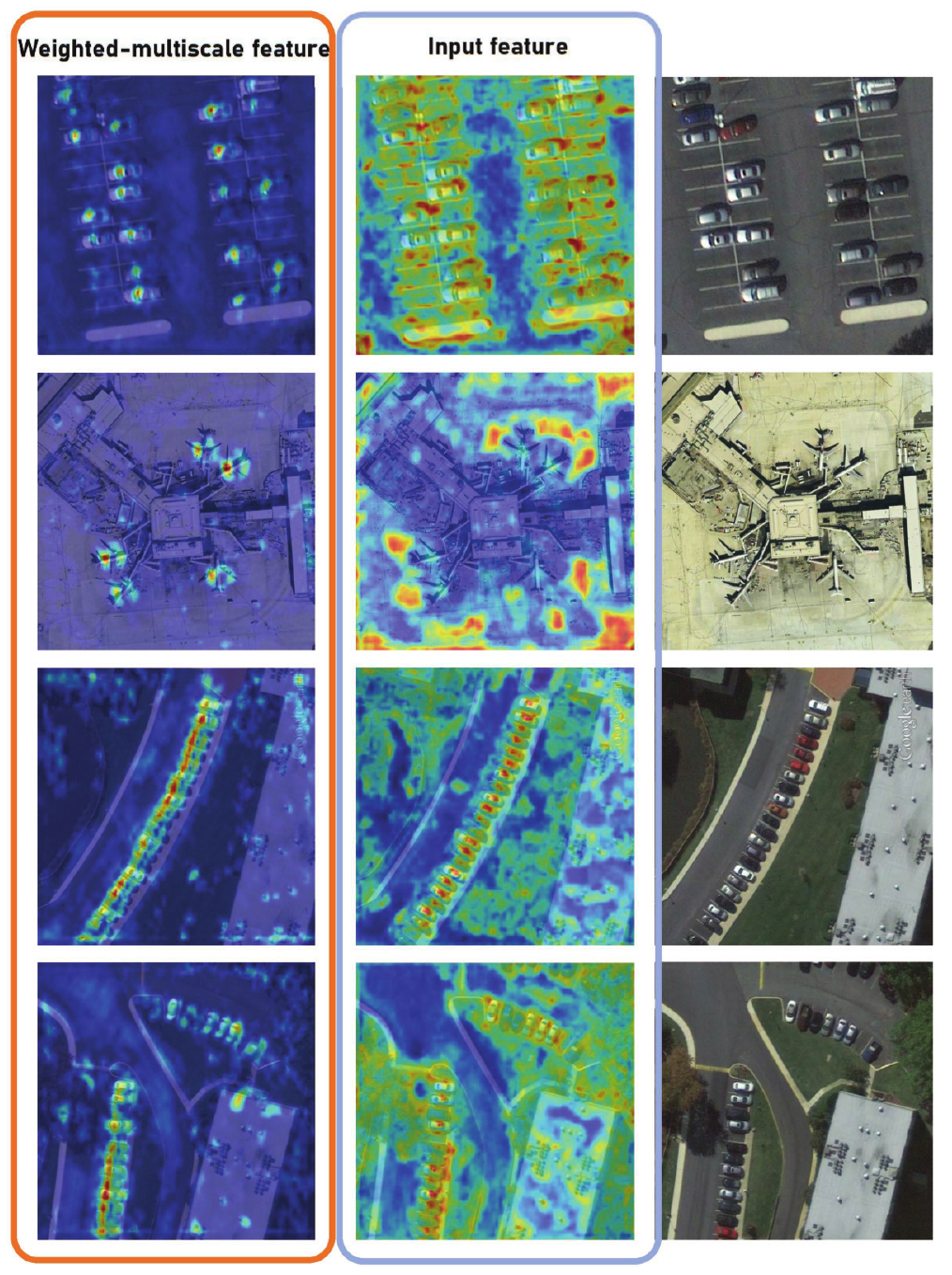}
			%\caption{fig2}
		\end{minipage}%
	}%
	\caption{(a) Structure diagram of convolution on weighted multi-scale extracting multi-scale object features while taking channel attention into account. (b) CAM for generating attention features by weighted multi-scale convolution.}\label{fig4}
\end{figure}

A pyramidal convolution is proposed in PyConv\cite{duta2020pyramidal}, which uses multiple scale convolution kernels to process the input features to solve the problems of multi-category capture and scale diversity. Multi-scale convolution brings more background noise when capturing finer object information, which is unfavorable to the extraction of object features in complex remote sensing images, for that reason, we improve the traditional multi-scale convolution by applying channel weighting to the features extracted under different Therefore, we improve the traditional multi-scale convolution by channel weighting the features extracted under different perceptual fields, paying attention to the importance of each layer of channels while extracting spatial features. Specifically, Figure \ref{fig4}a shows the weighted multi-scale convolution used to extract pixel level attention. After the number of channels of the fused features is changed through convolution, the feature information of multiple scales is extracted through m convolution kernels with different scales. The channel dimensions of each convolution output are $\frac{c}{m}$ input (with the  size 4 in this paper). The global adaptive average pooling is used to obtain 1$\times$1-size tensor after splicing these features. $\mathcal{W}$ is obtained by using full connection layer and active layer after shape transformation of 3-dimensional tensor.
\begin{equation}
	\mathcal{W}={\Gamma}({\rm gap}({\rm cat}(\mathcal{F}_0,\mathcal{F}_1,\cdots,\mathcal{F}_{m}))), \mathcal{F}_i\in\mathbb{R}^{\frac{c}{m}\times\frac{w}{4}\times\frac{h}{4}},i\in\left [ {0,m} \right ] 
\end{equation}
Where, $\Gamma$() represents the full connection layer mapping function,$\mathcal{W}$ gets $\mathcal{W}_i$ through linear layer and Softmax function, and get $\tilde{\mathcal{W}}$ after splicing, highlighting significant object information and suppressing noise information. 
\begin{equation}
	\mathcal{W}_{i,j} = \frac{e^{\Phi_{i}(\mathcal{W})_j}}{\sum_{k=0}^{\frac{c}{m}}e^{\Phi_{i}(\mathcal{W})_{k}}},\mathcal{W}\in\mathbb{R}^{1\times c},\mathcal{W}_i\in\mathbb{R}^{1\times\frac{c}{m}},j\in\left [ {0,\frac{c}{m}} \right ] 
\end{equation}
\begin{equation}
	\tilde{\mathcal{W}}={\rm{reshape}}({\rm cat}(\mathcal{W}_0,\mathcal{W}_1,\cdots,\mathcal{W}_{m})),\tilde{\mathcal{W}}\in\mathbb{R}^{{1}\times{1}\times{c}}
\end{equation}
Where, $\Phi_i()$ represents the mapping function of the i-th scale corresponding to the linear layer.Ultimately, it is used two convolutions to aggregate weighted classification and regression features respectively. Next, we embedded the module into the interactive double-branch up-sampling structure. The whole process is described as follows using the formula: 
\begin{equation}
	\tilde{\mathcal{F}_t}={{\rm conv}_t}(\tilde{\mathcal{W}}\otimes {{\rm {cat}}}(\mathcal{F}_0,\mathcal{F}_1,\cdots,\mathcal{F}_{m})),\tilde{\mathcal{F}_t}\in\mathbb{R}^{{c}\times\frac{w}{4}\times\frac{h}{4}},(t=0,1)
\end{equation}

We used Grade CAM to show the final effect in Figure \ref{fig4}b. It can be seen that the channel weighting method improves the difference between the object and the background, and suppresses background noise.  
\subsection{Representation of boundary box parameters and Loss function}\label{subsec5}
This paper describes OBB using bearing angle and two rays. For a object in an aviatic remote sensing image, firstly, it is obtained the center point coordinates, and calculated the offset under the output scale according to the center point coordinates $\left ( x_{\rm c},y_{\rm c} \right )$ to follow:
\begin{equation}
	\delta_{\rm x} =\frac{x_{\rm c}}{d}-\left[\frac{x_{\rm c}}{d}\right],\delta_{\rm y}=\frac{y_{\rm c}}{d} -\left [\frac{y_{\rm c}}{d}\right]
\end{equation}
Where, $\left [\right ]$ represents the downward rounding function. $d$ represents the down sampling multiple factor, taking the object center point $\left ( x_{\rm c},y_{\rm c} \right )$ as the pole. The positive half axis xis as the positive direction of the polar coordinate. We calculate the angle of the center point on four sides by selecting the long sideand its corresponding angle as well as the short side as the label information for the initial label file $ ( \varphi,R_1,R_2,\delta_{\rm x},\delta_{\rm y})$. Here, $\varphi\in(-\frac{\pi }{2},\frac{\pi }{2} )$ Notwithstanding the different definition method in long edge from the representation method proposed in this paper, so can it to avoid the interchange of length and width.

\textbf{Joint loss function}. The network is bound to predict $( \overline{\varphi},\overline{R_1},\overline{R_2},\overline{\delta_{\rm x}},\overline{\delta_{\rm y}})$ as the rotation bounding box. Initially, it is used the loss function  $SmoothL1$ to regress the length, width and offset of the bounding box:

\begin{equation}
	\mathcal{L}_{\rm reg}={\rm SmoothL1}(( \varphi,R_1,R_2,\delta_{\rm x},\delta_{\rm y}), (\overline{\varphi},\overline{R_1},\overline{R_2},\overline{\delta_{\rm x}},\overline{\delta_{\rm y}}))
\end{equation}

$( \varphi,R_1,R_2,\delta_x,\delta_y)$ represents the tag value of Ground Truth, with $(\overline{\varphi},\overline{R_1},\overline{R_2},\overline{\delta_{\rm x}},\overline{\delta_{\rm y}})$ representing the predicted value. We found that if the angle and length and width are regressed separately following research, the network does not learn the association information between the angle and side length. Thanks to the existence of the direction angle, the model is unstable in the regression process. Polar Mask proposed a polar plane based IoU calculation method for instance segmentation tasks\cite{xie2020polarmask}, on which we propose a joint learning loss function (JIoU). We, for the initial labeling file, found that it is extremely complicated in solving the IoU of two arbitrary directional bounding boxes in the Cartesian coordinate system. This is because a formula for calculating the intersection area that can cover all cases is not available in Cartesian coordinates on the one hand, and by modeling the bounding box in the two-dimensional plane, a differentiable function describing the intersection area cannot be found on the other. The rectangle fails to use a function in the Cartesian coordinate system to describe the relationship between points on the bounding box, which is not conducive to the calculation of the bounding box loss. In the process of research, we found that it is unique on the corresponding internal ellipse of the bounding box, with the parameter information of the ellipse capable of representing the oriented bounding box. The ellipse can be represented by the polar diameter and polar angle in polar coordinates, which is a function of one-to-one mapping. We approximate the rectangular bounding box into a polar elliptic equation using the following equation based on the relationship between the ellipse and the rectangle.

\begin{equation}
	\rho =\sqrt{\frac{R_{1}^{2}*R_{2}^{2}} {R_{2}^{2}*\cos(\theta -\varphi)+R_{1}^{2}*\sin(\theta -\varphi)}},x=\rho*\cos(\theta),y=\rho*\sin(\theta) 
\end{equation}

Where, $R_1$, $R_2$ represents the long side and short side of the ellipse respectively, corresponding to half of the length and width of the rectangular boundary box.  represents the rotation angle of the rectangular boundary box. For any rectangular boundary box following the above transformation, there exists a unique elliptic equation corresponding to it. The above elliptic equation needs to be discretized in order to enable the computer to handle the intersection and union ratio of two ellipses, as shown in Figure \ref{fig2}a. The polar coordinate plane is divided into n equal parts with the object center point as the origin, with the corresponding angle of each is $\frac{2\pi}{n}$ . As such, a polar coordinate plane is divided into  angles as shown below:
\begin{equation}
	\Theta ^{\rm d} =\left \{ \theta_0^{\rm d},\theta_1^{\rm d},\cdots,\theta_i^{\rm d} \right \}, \theta_i^{\rm d}=\frac{2*\pi*i}{n}(i=0,1,\cdots,n-1)
\end{equation}

The discrete polar coordinate equation of the ellipse can be obtained following the discretized polar coordinate plane angle. The calculation method is as follows:

\begin{equation}
	\rho_i^{\rm d} =\sqrt{\frac{R_{1}^{2}*R_{2}^{2}} {R_{2}^{2}*\cos(\theta_i^{\rm d}  -\varphi)+R_{1}^{2}*\sin(\theta_i^{\rm d}  -\varphi)}},x=\rho*\cos(\theta_i^{\rm d} ),y=\rho*\sin(\theta_i^{\rm d} ) 
\end{equation}

After obtaining the conversion formula of the polar coordinate equation between the boundary box and the discrete ellipse, we can get two intersecting discrete ellipses in keeping with the predicted value and label value. In Figure \ref{fig2}b, orange represents the discrete ellipse corresponding to Ground Truth, with orange representing the discrete ellipse corresponding to the prediction box. As reference to the calculation method of IoU, it is obtained on the intersection and union ratio of the areas of two discrete ellipses. In the figure, blue represents the intersection part, with red representing the union part. As per the area integral formula of the elliptic equation in polar coordinates, it is approximately calculated on the intersection area, union area and their ratio. The formula is described as follows:

\begin{equation}
	\overline{{\rm JIoU}} = \frac{ {\textstyle \sum_{i=0}^{n-1}*\frac{1}{2}*{\rm min}(\overline{\rho}_i^{\rm d},{\rho}_i^{\rm d})^2*\frac{2*\pi}{n}}}{{\textstyle \sum_{i=0}^{n-1}*\frac{1}{2}*{\rm max}(\overline{\rho}_i^{\rm d},{\rho}_i^{\rm d})^2*\frac{2*\pi}{n}}} 
\end{equation}

Where, $\overline{\rho}_i^{\rm d}$ and $\rho_i^{\rm d}$ respectively represent discrete ellipse prediction values and label values at corresponding angles. We will further explain the nature of JIoU through experiments in the \ref{subsec9} chapter. Finally, the above formula is further simplified to obtain: 

\begin{equation}
	\overline{{\rm JIoU}} = \frac{ {\textstyle \sum_{i=0}^{n-1}*\min(\overline{\rho}_i^{\rm d},{\rho}_i^{\rm d})^2}}{{\textstyle \sum_{i=0}^{n-1}*\max(\overline{\rho}_i^{\rm d},{\rho}_i^{\rm d})^2}} 
\end{equation}

In the process of back propagation, the smaller the loss value, the better. As such, the actual JIoU is:

\begin{equation}
	{\rm JIoU} =  \mathbb{E}(-\log(\overline{{\rm JIoU}}))
\end{equation}

The ${\rm JIoU}$ proposed in this paper has the following characteristics: Initially, the loss function is used as a tool for optimizing parameters, and ${\rm JIoU}$ is differentiable, which supports back propagation. Subsequently, ${\rm JIoU}$ can solve the periodic problem in the angle regression process. Compared with using SmoothL1 as the loss function of the bounding box regression directly, ${\rm JIoU}$ improves the correlation between the angle and the side length of the bounding box, thus effectively improving the accuracy of the bounding box parameter regression.

\textbf{For other loss functions}. The accurate prediction of poles is very important for the positioning of the bounding box. With reference to the representation method of the Heatmap in CenterNet, this paper uses the Gaussian kernel function $h(x,y) =e^{-\frac{(x-x_{\rm p})^2+(y-y_{\rm p})^2}{2*\sigma^2} } $ to fill the center point in the Heatmap. In aviatic remote sensing images, it is easy to cause the imbalance of positive and negative samples owing to large differences in object scales, small object density and other problems. As such, for the optimization of Heatmap, it, with Focal loss, is calculated on the loss value of positive samples corresponding to each object under different scales.
\begin{equation}
	\mathcal{L}_{{\rm cla}}=-\frac{1}{N}*\sum\begin{cases}(1-pt)^\gamma *{\rm log}(pt),y=1\\ (1-y)^\alpha *pt^\gamma *{\rm log}(1-pt),others\end{cases}
\end{equation}
Where, $N$ represents the number of positive samples filtered corresponding to the object in the input image, which is used to normalize the loss value. $\alpha$ and $\gamma$ represent the super parameter of Focal loss. The size set in this paper is 4 and 2. Each object value corresponds to a positive sample of network prediction. The total network loss obtained from the above tasks is:
\begin{equation}
	\mathcal{L}=\mathcal{L}_{{\rm cla}}+\mu*{\rm JIoU}+\mathcal{L}_{{\rm reg}}
\end{equation}
Indicates the weight coefficient to reduce the balance loss, set to 5.0.

\textbf{Inference process}. Since the boundary box in object detection is drawn in a rectangular coordinate system, the predicted boundary box parameters need to be decoded. The specific process is as follows: It is calculated the horizontal boundary box coordinates from the two rays in the prediction results:

\begin{equation}
	\begin{bmatrix}
		\overline{x}_i\\\overline{y}_i
	\end{bmatrix}=\begin{bmatrix}
		\sqrt{2}*\sin(\frac{i*\pi}{2}-\frac{\pi}{4} )*\overline{R}
		_1\\\sqrt{2}*\sin(\frac{(i-1)*\pi}{2}-\frac{\pi}{4} )*\overline{R}_2
	\end{bmatrix},(i=0,1,2,3)
\end{equation}

Where,$(x_i,y_i)$ represents the four corners of the horizontal bounding box, with the calculation direction in clockwise order. It is used the rotation angle in the prediction result to construct a clockwise rotation transformation matrix, and multiplied the matrix and the horizontal bounding box coordinates to obtain the rotation coordinates to follow. After that, it is extracted the center point coordinates $({x_{\rm c},y_{\rm c}})$ of the object according to the predicted Heatmap and Offset, and then added the rotation coordinates and the object center point coordinates to obtain the final rotation coordinates:

\begin{equation}
	\begin{bmatrix}
		x_i^{\rm r}\\y_i^{\rm r}
	\end{bmatrix}=\begin{bmatrix}
		\cos(\overline{\varphi})&-\sin(\overline{\varphi}) \\
		\sin(\overline{\varphi}) &\cos(\overline{\varphi})
	\end{bmatrix}\cdot \begin{bmatrix}
		\overline{x}_i\\\overline{y}_i
	\end{bmatrix}+\begin{bmatrix}
		x_{\rm c}\\y_{\rm c}
	\end{bmatrix}
\end{equation}
\section{Test and Experiment}\label{sec4}
In this section, experiments are designed to verify the detection performance of BWP-Det, compared with other detectors of the same type.
\subsection{Dataset}\label{subsec6}
The proposed model will be tested on NWPU VHR-10, UCAS-AOD and DOTA, three challenging remote sensing data sets. These three data sets contain a large number of directional objects and horizontal objects, with these pictures taken by UAVs and satellites.

DOTA represents the largest dataset in the detecting task to aviatic remote sensing object, which contains 2806 aviatic remote sensing images. It has two annotation methods: directional bounding box and horizontal bounding box. In this experiment, it is used only on the labels labeled by the directional bounding box. In this data set, the ratio of training set, verification set and test set is 1/2, 1/6 and 1/3 respectively.

NWPU VHR-10 is a detecting dataset with 10 categories to airborne remote sensing object. The dataset contains 800 ultra-high resolution remote sensing images, including 650 foreground images and 150 background images. During the experiment, the training set and test set were randomly divided by 8:2. NWPU VHR-10 dataset has only horizontally labeled labels. It is used this dataset to verify the performance of the model in horizontal detection tasks.

The UCAS-AOD dataset contains a total of 2420 pictures and 14596 objects. UCAS-AOD includes two categories: aircraft and vehicles, including 7482 and aircraft objects in 1000 images and 7114 vehicle objects in 510 images. The objects in UCAS-AOD have two labeling methods: horizontal bounding box and directional bounding box. In this experiment, we randomly selected 1110 and 400 for training and testing respectively as per\cite{zhao2021polardet}.
\subsection{Train and Test Details}\label{subsec7}
This experiment was carried out on a single RTX 3090 with a 24g video memory size. In the training phase, the shape of the input images of BWP-Det remains 608 × 608 size square in order to prevent the object from deforming due to the change of image size.

For DOTA dataset, the pixel size range of the image represens $800\times800\sim4000\times4000$. The original image is cropped to the size of 1024$\times$1024 at a repetition rate of 0.25 after scaling [0.5, 1] in order to ensure the integrity of the object. The processing method of UCAS-AOD dataset is the same as\cite{zhao2021polardet}, with the processing method of NWPU VHR-10 dataset the same as\cite{zhou2020arbitrary}. Adam was used as the optimizer in the model. The momentum is set to 0.9, with the initial learning rate 0.000125. The cos descent method is adopted, with the batch size set to 16.The three datasets train for 150, 100 and 50 epochs. In the reasoning stage, it was the large scale of the test pictures in DOTA that empower pictures to be cut by the same method as training. The reasoning results are integrated using the toolbox provided by the DOTA official website, evaluated by the official website server. The same method as that used in training was adopted for tailoring and integration for UCAS-AOD and NWPU VHR-10. AP score was used as the evaluation index of the model. The calculation method in Pascal Visual was adopted when calculating AP, with IoU set to the default size of 0.5. In addition, it was also used some conventional data enhancement methods in this paper, including image radial transformation, random horizontal up and down flip transformation and color enhancement.
\subsection{Comparison with state-of-the-art detectors}\label{subsec8}
\begin{figure}[htbp]
	\centering
	\subfigure[]{
		\begin{minipage}[t]{0.3\linewidth}
			\centering
			\includegraphics[width=1.5in]{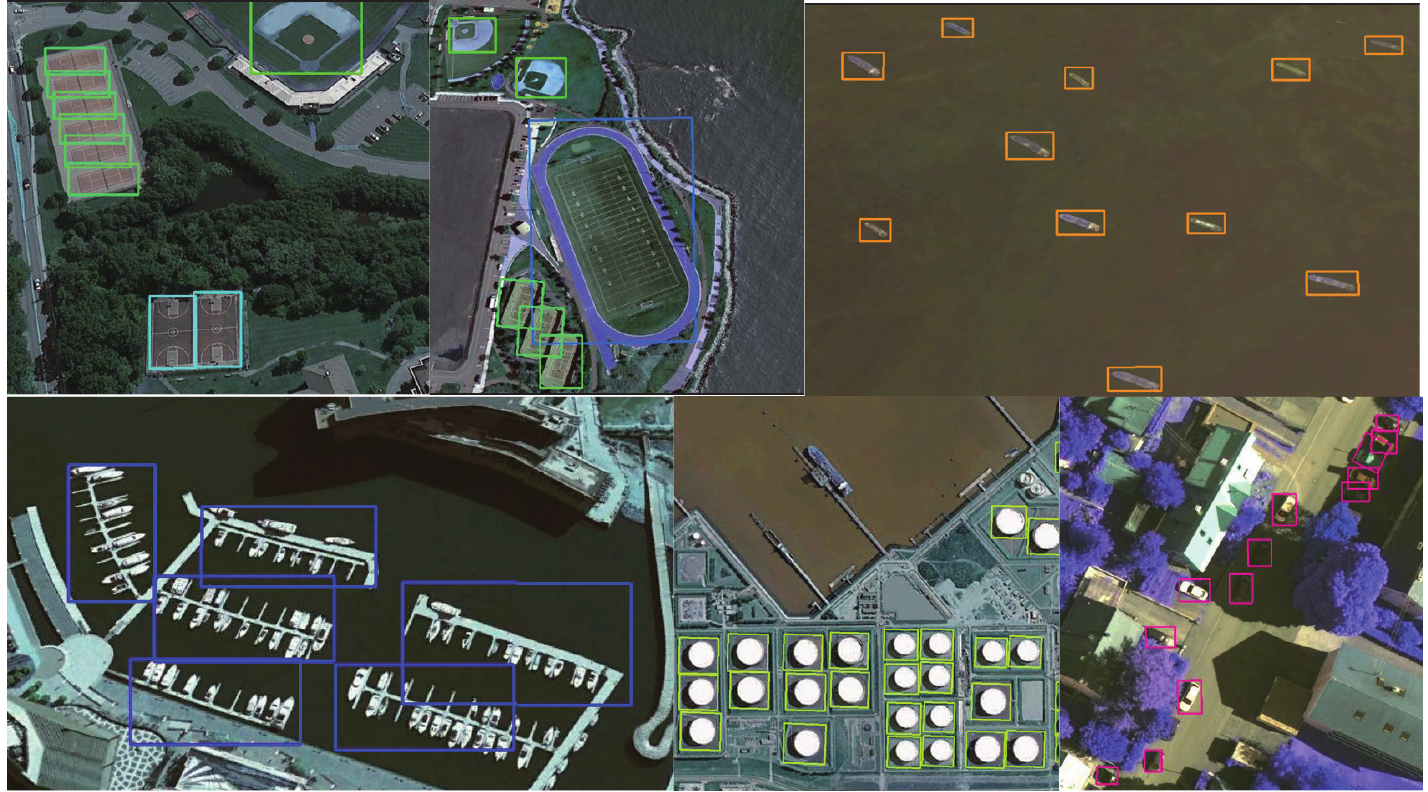}
			%\caption{fig1}
		\end{minipage}%
	}%
	\subfigure[]{
		\begin{minipage}[t]{0.4\linewidth}
			\centering
			\includegraphics[width=1.3in]{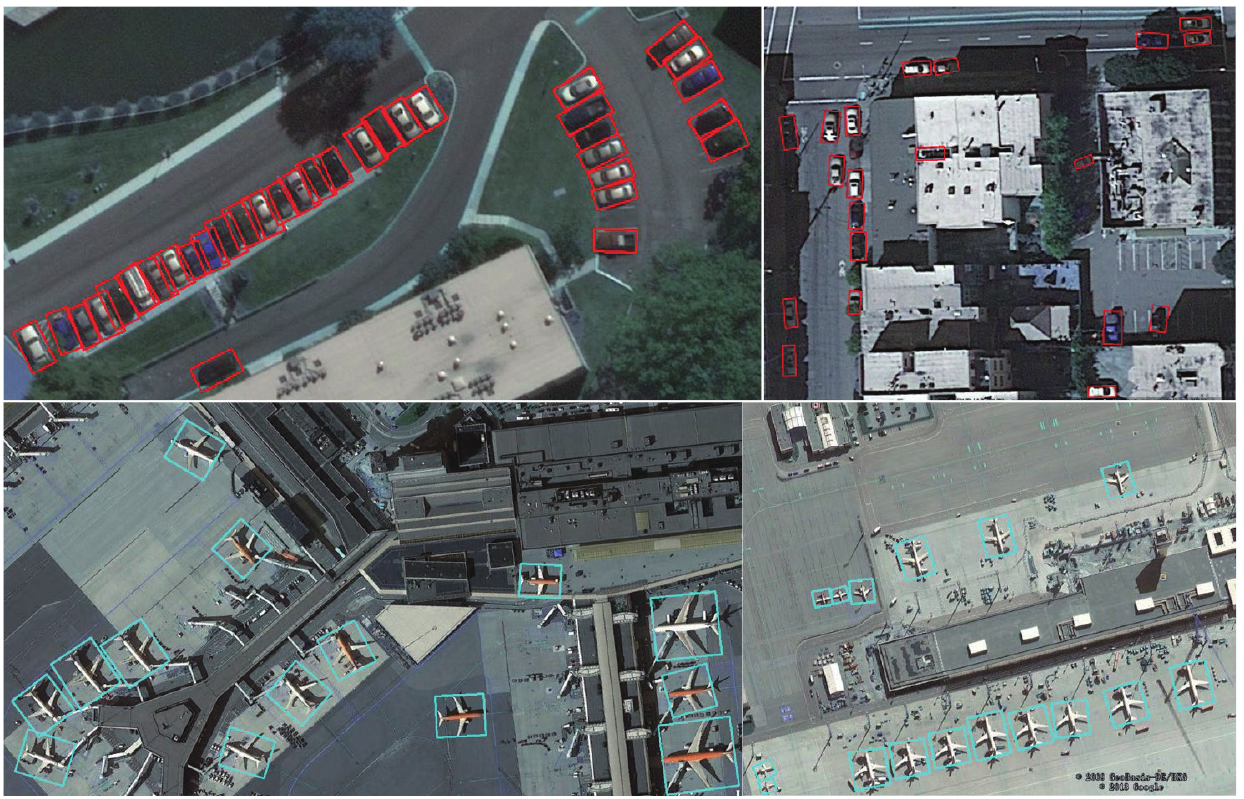}
			%\caption{fig2}
		\end{minipage}%
	}%
	\subfigure[]{
		\begin{minipage}[t]{0.25\linewidth}
			\centering
			\includegraphics[width=1.6in]{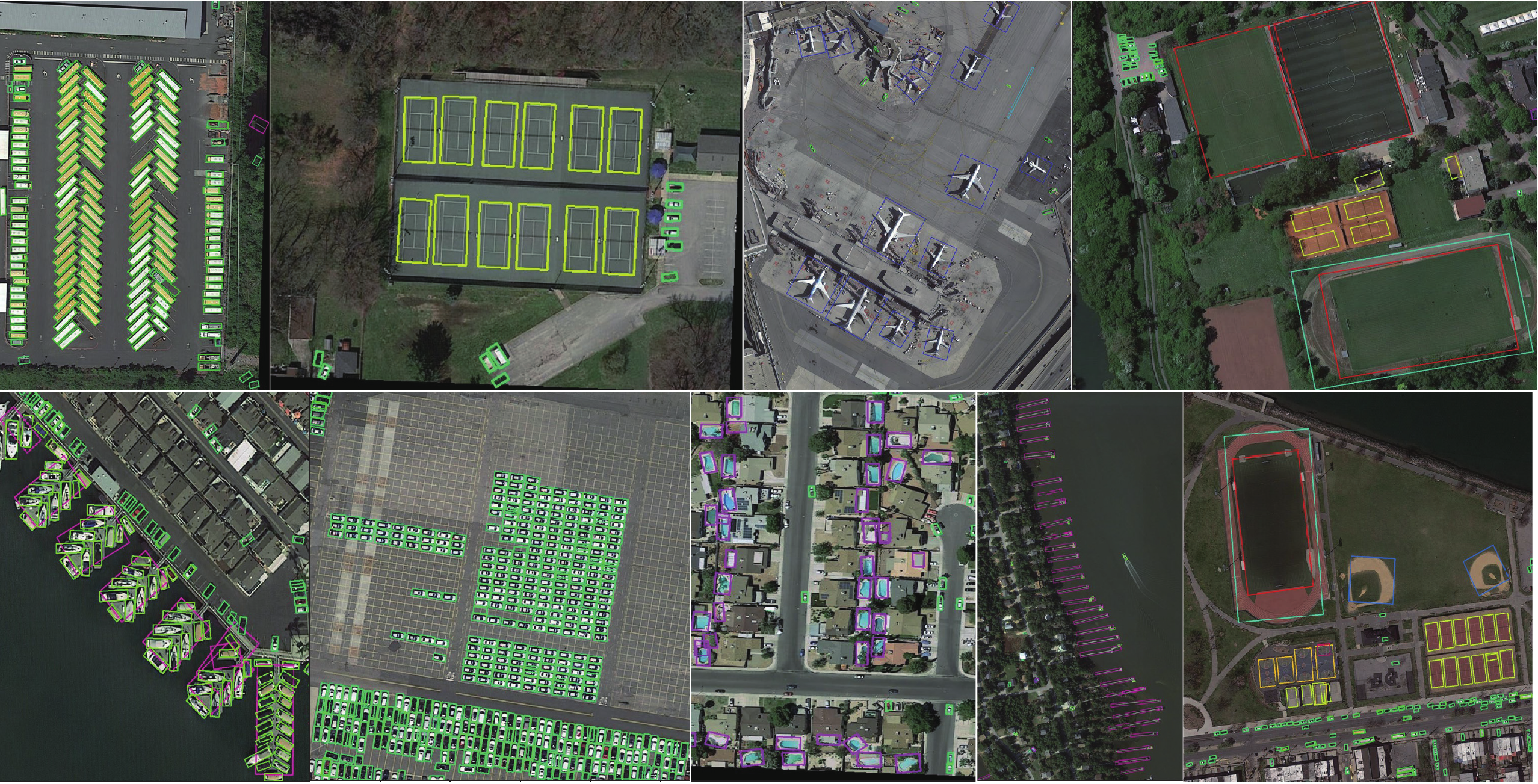}
			%\caption{fig2}
		\end{minipage}%
	}%
	\centering
	\caption{BWP-Det test results in NWPU VHR-10 (a), UCAS-AOD (b)and DOTA datasets (c).}\label{fig6}
\end{figure}

The validity of the models was conducted on NWPU VHR-10, UCAS-AOD and DOTA datasets respectively in this paper, and show the results in the Figure \ref{fig6}.Table \ref{tab1} shows the comparison results of BWP-Det and other models on NWPU VHR-10 dataset HBB tasks. The best result is 92.07$\%$ when Backbone remained ResNext50. Highly good results have also been achieved on the UCAS-AOD dataset. As shown in Table \ref{tab2}, the mAP has reached 97.88$\%$. Table \ref{tab3} shows the results of the model on the DOTA dataset. The test results were processed using the tools provided by DOTA, and submitted to the official website server for evaluation to follow. As the input scale remained 608 × Under 608, the mAP reached 71.42$\%$. Despite the fact that this result is inferior to other models, we can see that ResNext50 is selected as the Backbone in this test. Furthermore, our model is the structure of Anchor Free, which has more advantages in model parameters and detection speed. The ablation experiment also verified this conclusion. These results are more competitive than the current models of the same type, having reached SOTA on some datasets. However, it is simpler in the model design proposed in this paper, with no more super parameters needed to be set.
\begin{table}[h]
	\centering
	\caption{mAP results of NWPU VHR-10 dataset ($\%$). }\label{tab1}%
	\scalebox{0.75}{
		\begin{tabular}{@{}lccccccccccc@{}}
			\toprule
			Models &ap&sh&st&bd&tc&bc&gtf&hb&br&ve&mAP\\
			\midrule
			P-RSDet\cite{zhou2020arbitrary}    &97.90&	92.40&	88.30&	95.80&	89.30&	96.20&	94.90&	81.90&	83.30&	88.70&	90.80 \\
			Faster R-CNN\cite{7485869}    & 99.28&	96.67&	72.51&	98.76&	79.91&	37.95&	88.04&	72.64&	62.29&	97.23&	80.52  \\
			SSD\cite{zhou2020arbitrary}    &96.40&	87.80&	84.10&	93.60&	89.60&	92.50&	95.70&	81.20&	79.20&	83.90&	88.60 \\
			Centernet\cite{zhou2019objects}+JIoU & 90.40&	60.90&	79.80&	89.90&	82.60&	80.60&	98.30&	73.40&	76.70&	53.10&	78.40  \\
			Ours     & 99.25&	92.85&	96.47&	97.42&	95.37&	96.42&	99.31&	86.37&	74.50&	82.73&	92.07 \\
			\bottomrule
	\end{tabular}}
\end{table}
\begin{table}[h]
	\centering
	\caption{mAP results of UCAS-AOD dataset ($\%$). }\label{tab2}%
	\begin{tabular}{lccc}
		\toprule
		Models & Plane & Car & mAP \\ 
		\midrule
		YOLOv2\cite{redmon2017yolo9000}  & 96.60 & 79.20 & 87.90 \\ 
		R-DFPN\cite{yang2018automatic} & 95.90 & 82.50 & 89.20 \\ 
		DRBox\cite{liu2017learning} & 94.90 & 85.00 & 89.95 \\ 
		S2ARN\cite{bao2019single}& 97.60 & 92.20 & 94.90 \\ 
		RetinaNet-H \cite{yang2019scrdet} & 97.34 & 93.60 & 95.47 \\ 
		ICN\cite{azimi2019towards} & - & - & 95.67 \\
		FADet\cite{li2019feature}& 98.69 & 92.72 & 95.71 \\
		R3Det\cite{yang2021r3det}& 98.20 & 94.14 & 96.17 \\
		SCRDet++\cite{yang2022scrdet++} & 98.93 & 94.97 & 96.95 \\
		PolarDet\cite{zhao2021polardet}  & 99.08 & 94.96 & 97.02 \\
		Ours (BWP-Det) & 98.77 & 97.00 & 97.88 \\ 
		\bottomrule
	\end{tabular}
\end{table}
\begin{table}[h]
	\centering
	\begin{minipage}{\textheight}
		\caption{mAP results of DOTA dataset ($\%$).}\label{tab3}
		\scalebox{0.47}{
			\begin{tabular}{lccccccccccccccccc}
				\toprule
				Models&Backbone & Pl & Bd & Br & Gtf & Sv & Lv & Sh & Tc & Bc & St & Sbf & Ra & Ha & Sp & He & mAP \\ \hline
				\textbf{Anchor Based} &~& ~ & ~ & ~ & ~ & ~ & ~ & ~ & ~ & ~ & ~ & ~ & ~ & ~ & ~ & ~ & ~ \\ \hline
				R2CNN\cite{jiang2017r2cnn} &ResNet101& 80.94 & 65.67 & 35.34 & 67.44 & 59.92 & 50.91 & 55.81 & 90.67 & 66.92 & 72.39 & 55.06 & 52.23 & 55.14 & 53.35 & 48.22 & 60.67 \\ 
				RRPN\cite{ma2018arbitrary}&ResNet101 & 88.52 & 71.20 & 31.66 & 59.30 & 51.85 & 56.19 & 57.25 & 90.81 & 72.84 & 67.38 & 56.69 & 52.84 & 53.08 & 51.94 & 53.58 & 61.01 \\ 
				ICN\cite{ma2018arbitrary}&ResNet101& 81.40 & 74.30 & 47.70 & 70.30 & 64.90 & 67.80 & 70.00 & 90.80 & 79.10 & 78.20 & 53.60 & 62.90 & 67.00 & 64.20 & 50.20 & 68.20 \\ 
				RetinaNet\cite{zhao2021polardet}&ResNet101 & 88.92 & 67.67 & 33.55 & 56.83 & 66.11 & 73.28 & 75.24 & 90.87 & 73.95 & 75.07 & 43.77 & 56.72 & 51.05 & 55.86 & 21.46 & 62.02 \\ 
				RoI Trans\cite{ding2019learning}&ResNet101 & 88.64 & 78.52 & 43.44 & 75.92 & 68.81 & 73.68 & 83.59 & 90.74 & 77.27 & 81.46 & 58.39 & 53.54 & 62.83 & 58.93 & 47.67 & 69.56 \\ 
				Gliding Vertex\cite{xu2020gliding}&ResNet101 & 89.64 & 85.00 & 52.26 & 77.34 & 73.01 & 73.14 & 86.82 & 90.74 & 79.02 & 86.81 & 59.55 & 70.91 & 72.94 & 70.86 & 57.32 & 75.02 \\ 
				R3Det\cite{yang2021r3det}&ResNet152& 89.24 & 80.81 & 51.11 & 65.62 & 70.67 & 76.03 & 78.32 & 90.83 & 84.89 & 84.42 & 65.10 & 57.18 & 68.10 & 68.98 & 60.88 & 72.81 \\ 
				SCRDet++(R3Det)\cite{yang2022scrdet++}&ResNet101 &89.20 & 83.36 & 50.92 & 68.17 & 71.61 & 80.23 & 78.53 & 90.83 & 86.09 & 84.04 & 65.93 & 60.8 & 68.83 & 71.31 & 66.24&74.41 \\ 
				SCRDet\cite{yang2019scrdet}&ResNet101& 89.98 & 80.65 & 52.09 & 68.36 & 68.36 & 60.32 & 72.41 & 90.85 & 87.94 & 86.86 & 65.02 & 66.68 & 66.25 & 68.24 & 65.21 & 72.61 \\ 
				RSDet\cite{qian2021learning}&ResNet152 & 90.1 & 82.0 & 53.8 & 68.5 & 70.2 & 78.7 & 73.6 & 91.2 & 87.1 & 84.7 & 64.3 & 68.2 & 68.1 & 69.3 & 63.7 & 74.1 \\\hline 
				\textbf{Anchor Free} & ~ & ~ & ~ & ~ & ~ & ~ & ~ & ~ & ~ & ~ & ~ & ~ & ~ & ~ & ~ & ~& ~ \\ \hline
				Centernet\cite{zhou2019objects}&ResNet101 & 81.00 & 64.00 & 22.60 & 56.60 & 38.60 & 64.00 & 64.90 & 90.80 & 78.00 & 72.50 & 44.00 & 41.10 & 55.50 & 55.00 & 57.40 & 59.10 \\ 
				TOSO\cite{feng2020toso}&ResNet101 & 80.17& 65.59& 39.82& 39.95& 49.71& 65.01& 53.58& 81.45 &44.66& 78.51& 48.85& 56.73& 64.40& 64.24& 36.75&57.92\\
				IENet\cite{lin2019ienet}&Res101 & 88.15& 71.38& 34.26 &51.78& 63.78 &65.63& 71.61& 90.11& 71.07& 73.63 &37.62& 41.52& 48.07& 60.53& 49.53&61.24\\
				PIoU\cite{chen2020piou}&DLA34 & 80.90 &69.70& 24.10 &60.20 &38.30& 64.40& 64.80&90.90&77.20 &70.40& 46.50 &37.10& 57.10& 61.90 &64.00&60.50\\
				BBAVectors\cite{yi2021oriented}&ResNet101 & 88.35 & 79.96 & 50.69 & 62.18 & 78.43 & 78.98 & 87.94 & 90.85 & 83.58 & 84.35 & 54.13 & 60.24 & 65.22 & 64.28 & 55.70 & 72.32 \\ 
				P-RSDet\cite{zhou2019objects}&ResNet101 & 88.58 & 77.84 & 50.44 & 69.29 & 71.10 & 75.79 & 78.66 & 90.88 & 80.10 & 81.71 & 57.92 & 63.03 & 66.30 & 69.77 & 63.13 & 72.30 \\ 
				O2-DNet\cite{wei2020oriented}&Hourglass104\cite{newell2016stacked} & 89.31 & 82.14 & 47.33 & 61.21 & 71.32 & 74.03 & 78.62 & 90.76 & 82.23 & 81.36 & 60.93 & 60.17 & 58.21 & 66.98 & 61.03 & 71.04 \\ 
				DRN\cite{pan2020dynamic}&Hourglass104\cite{newell2016stacked} & 88.91& 80.22& 43.52& 63.35& 73.48& 70.69& 84.94 &90.14&83.85&84.11& 50.12& 58.41& 67.62 &68.60& 52.50&70.70 \\
				EARL\cite{guan2023earl}&ResNet50 &89.76& 78.79 &47.01& 65.20 &80.98 &79.99& 87.33& 90.74& 79.17& 86.23& 49.09& 65.87& 65.75& 71.86& 55.21&72.87\\				
				Ours(BWP-Det)          &ResNext50&89.36&78.91& 51.06 & 65.34 & 61.41 & 77.35 & 76.23 & 89.34 & 85.36 & 82.00 & 57.22 & 61.60 & 65.67&64.23& 67.34& 71.42 \\ \bottomrule
		\end{tabular}}
	\end{minipage}
\end{table}

\subsection{Ablation Studies}\label{subsec9}
In this section, the ablation experiment results of four different strategies are presented, including different Backbone, JIoU loss functions, WmConv modules and IDB.
\begin{figure}[htbp]
	\centering
	\subfigure[]{
		\begin{minipage}[t]{0.2\linewidth}
			\centering
			\includegraphics[width=1.2in]{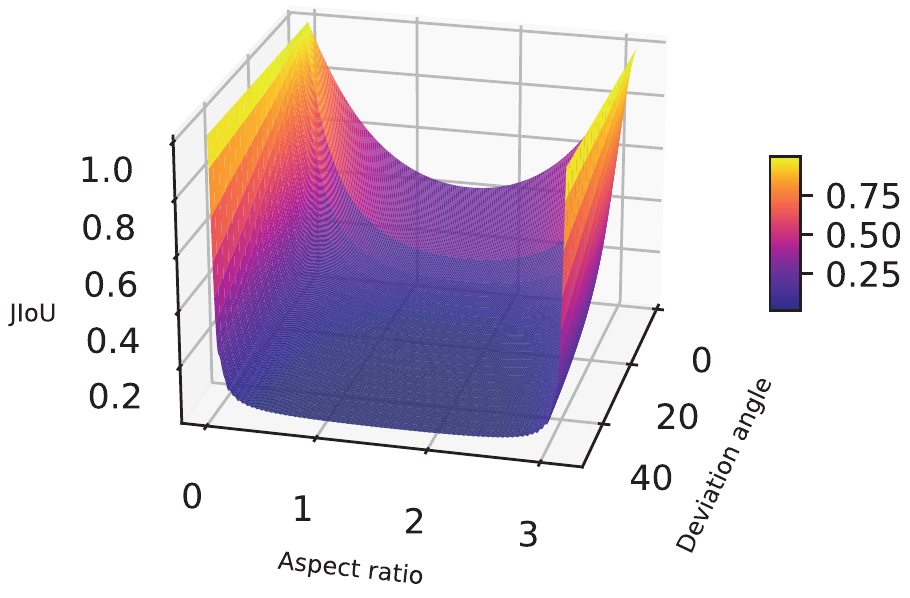}
		\end{minipage}%
	}%
	\subfigure[]{
		\begin{minipage}[t]{0.9\linewidth}
			\centering
			\includegraphics[width=3.5in]{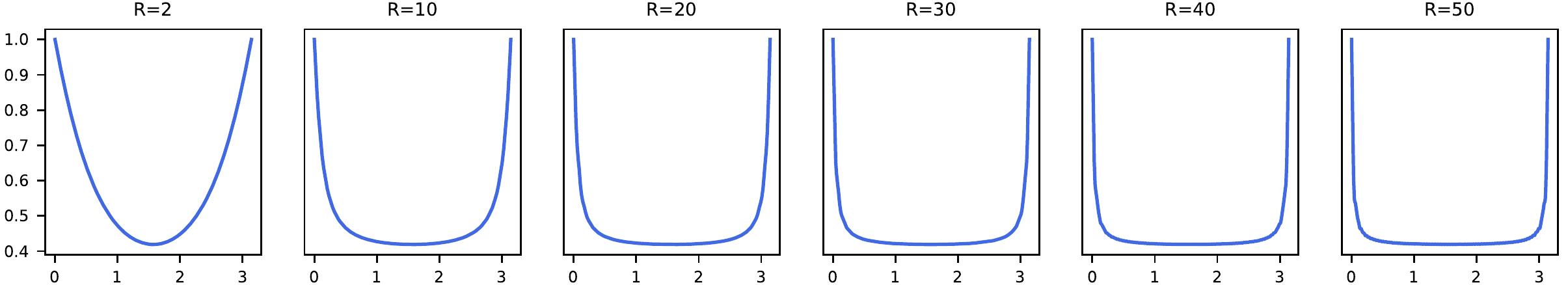}
			%\caption{fig2}
		\end{minipage}%
	}%
	
	\subfigure[]{
		\begin{minipage}[t]{0.2\linewidth}
			\centering
			\includegraphics[width=1.2in]{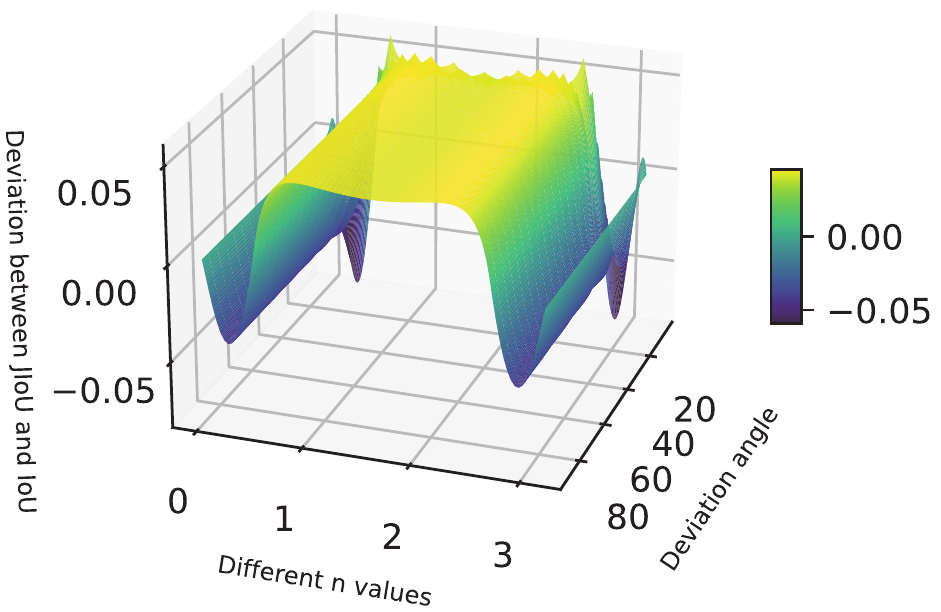}
			%\caption{fig2}
		\end{minipage}%
	}%
	\subfigure[]{
		\begin{minipage}[t]{0.9\linewidth}
			\centering
			\includegraphics[width=3.5in]{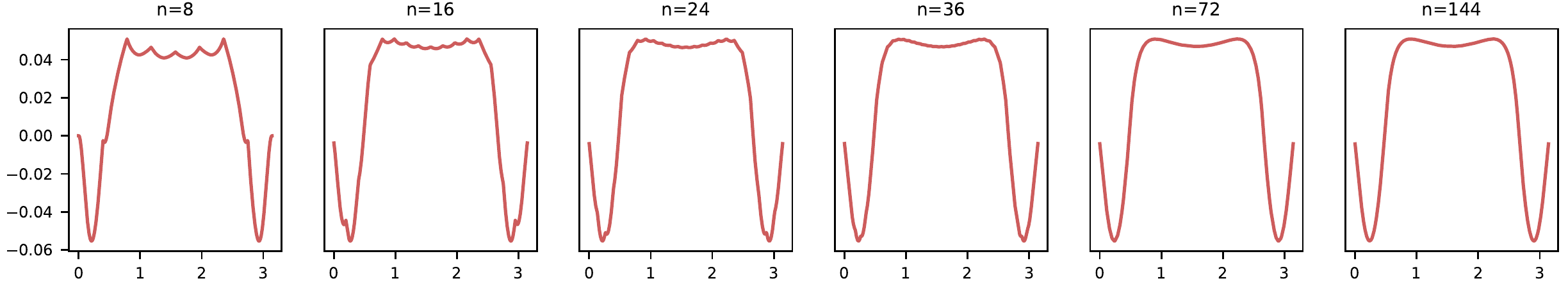}
			%\caption{fig2}
		\end{minipage}%
	}%
	\centering
	\caption{(a) For different n values, the relationship between the deviation between traditional IoU and JIoU and different angle differences. (b)  Deviation curves for different aspect ratios. (c)For different object aspect ratios, the relationship between JIoU and different angle differences. (d)  Deviation curves for different n values.}\label{fig7}
\end{figure}

\textbf{Backbone:} ResNext50 is used as the feature extraction network in this paper. In order to test the impact of different Backbones on the experimental results, ResNet18 is used as the feature extraction network instead of ResNext50 under the same other conditions, with experiments conducted on the UCAS-AOD directional object dataset. As shown in Table \ref{tab4}, the mAP of the model can still reach 96.37$\%$ of the final result after replacing Backbone with ResNet18. However, it is worth noting that the ResNet18 FPS tested under GTX 1050Ti has reached 46.29. Compared with ResNext50, the reasoning speed of the model has increased by approximately four times, which has reached the effect of real-time detection. The final results of the experiment also show that the model proposed in this paper is still applicable at the time of selected different Backbones.

\textbf{JIoU:} In the previous chapter, the loss function JIoU propose in this paper is described. To verify the effectiveness of this loss function, the UCAS-AOD directional object dataset is compared with the traditional SmoothL1 loss function. From the results in Table \ref{tab4}, it can be seen that after replacing the loss function with JIoU, the model's mAP results improve 5.03$\%$.

\begin{figure}[htbp]
	\centering
	\subfigure[]{
		\begin{minipage}[t]{0.5\linewidth}
			\centering
			\includegraphics[width=2.5in]{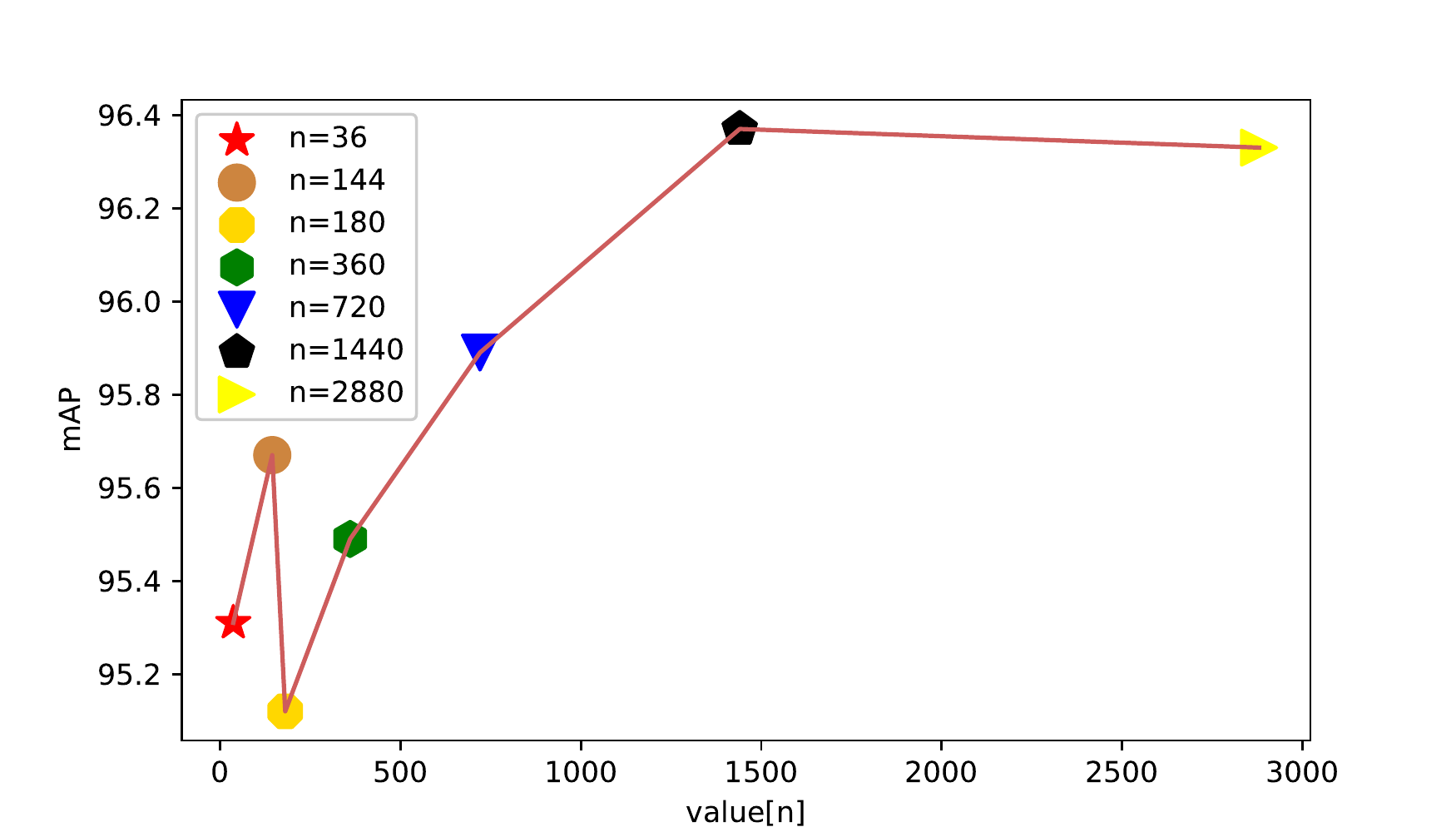}
		\end{minipage}%
	}%
	\subfigure[]{
		\begin{minipage}[t]{0.5\linewidth}
			\centering
			\includegraphics[width=2.5in]{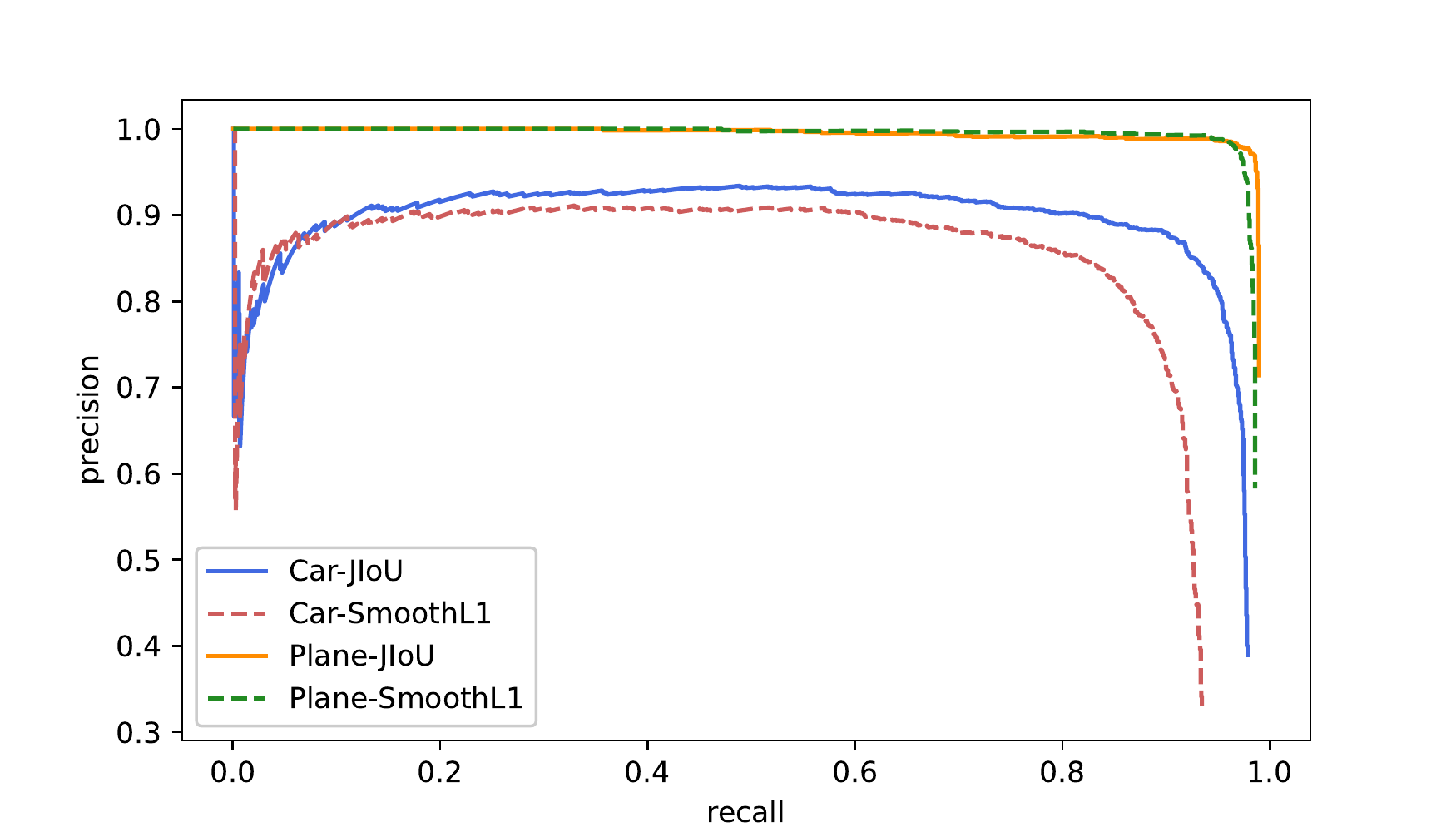}
			%\caption{fig2}
		\end{minipage}%
	}%
	\centering
	\caption{(a) Test results of different n values. (b)  PR curve of JIoU and smoothl1 }\label{fig8}
\end{figure}

Figure \ref{fig8}b shows the PR curves of JIoU and SmoothL1, It can also be seen from the figure that after using JIoU, the model recall rate has significantly improved, which shows that the JIoU proposed in this paper as a loss function has better performance in directional object detection tasks. From the accuracy of a single object, it can be seen that JIoU has significantly improved the detection accuracy of cars. From the detection effect of the boundary box in the results, two reasons can be found: first, although cars are smaller than aircraft, However, its aspect ratio is larger, which results in more sensitive results to angle changes when calculating IoU. Secondly, the angle has periodicity and the unit of expression is different from the length and width. In case of traditional SmoothL1, it is inconsistent in the optimization rate of angle and length and width. That leads to excessive optimization of local parameters of the model, ultimately leading to poor generalization of the model. However, JIoU calculates the extreme value of discrete ellipse through angle and jointly optimizes it, which effectively solves the above problems caused by angle, allowing the model to possess better generalization performance.

The sensitivity of the calculation method of JIoU to the aspect ratio of different objects is different. When the aspect ratio is small, the ellipse tends to be circular. At this time, the sensitivity of JIoU to the boundary box parameters will decrease, which can be seen in Figure \ref{fig7}ab. Figure \ref{fig7}cd shows the influence of different n values on JIoU. When the value of n increases gradually, the range of the deviation curve between traditional IoU and JIoU becomes smaller and smaller. At this time, the accuracy of the calculation of IoU is further improved.
\begin{algorithm}[t]
	\caption{JIoU}
	\hspace*{0.02in} {\bf Input:}
	input $In_p$,$In_t$,input shape as [3, - 1];n=720\\
	\hspace*{0.02in} {\bf Output:}
	Loss
	\begin{algorithmic}[1]
		\For{i$ \gets $0 to n-1 }
		\State $\Theta^{\rm d}[i] \gets  \frac{2*\pi*i}{n}$
		\EndFor
		\For{c to  p,t}
		\State $\rho_{[c]}^{\rm d}\gets\sqrt{\frac{In[1]_{[c]}^{2}*In[2]_{[c]}^{2}} {In[2]_{[c]}^{2}*\cos(\Theta^{\rm d}  -In[0]_{[c]})+In[1]_{[c]}^{2}*\sin(\Theta^{\rm d}  -In[0]_{[c]})}}$
		\EndFor
		\State Loss$\gets -{\rm log(\frac{{\rm sum({\rm min(\rho_{[p]}^{\rm d},\rho_{[t]}^{\rm d})})}}{{\rm sum({\rm max(\rho_{[p]}^{\rm d},\rho_{[t]}^{\rm d})})}})} $
		\State \Return Loss
	\end{algorithmic}
\end{algorithm}

At the same time, we use different n values to train on the UCAS-AOD data set, and test. As can be seen from Figure \ref{fig8}a, when $n$ values are small, the test results show randomness. With the increase of n, the accuracy also gradually increases, and the improvement is also smaller and smaller, and finally tends to be stable, showing a nonlinear relationship as a whole.

\textbf{Interactive double-branch:} IDB represens a method of reasonably allocating and fusing features through decoupling. It captures texture information and semantic information of delicacy by fusing middle layer features, gradually sampling classification and regression in the sub network to separate task related features and unrelated features. In Table \ref{tab4}, we compare the results of Dual Dual and Baseline under the same conditions. It can be seen that, compared with Baseline, the mAP of the model has reached 97.01$\%$ as using IDB. Besides, it can be found that the FPS of IDB decreases slightly compared with Baseline.
\begin{table}[h]
	\centering
	\caption{Baseline results in the first row, compared the AP($\%$) and mAP ($\%$) using different strategies in the baseline method. }\label{tab4}%
	\scalebox{0.83}{
		\begin{threeparttable}
			\begin{tabular}{ccccccccc}
				\toprule
				\multicolumn{2}{c}{Method} & Backbone &FLOPs(GMac) & Params(M) & FPS\tnote{1} & Plane&Car&mAP\\ 
				\hline
				Loss Function  & Models & ~ &  ~ & ~ & ~ & ~ & ~ &~ \\ \hline
				SmoothL1 & Centernet & ResNext50 & -& -& - &  98.23& 82.45 & 90.33 \\
				JIoU & Centernet & ResNext50 & 53.08 & 34.22 & 20.10 & 98.60 & 92.88 & 95.38 \\ 
				JIoU & Ours\tnote{*} & ResNext50 & 66.44  & 18.52  & 15.91 & 98.54 & 95.47 & 97.01 \\ 
				JIoU &Ours\tnote{+} & ResNet18 & 13.25  & 3.85  & 46.29 & 98.59 & 94.15 & 96.37 \\ 
				JIoU & Ours\tnote{+} & ResNext50 & 70.09  & 19.19  & 13.83 & 98.78 & 97.00 & 97.89 \\ 
				\bottomrule
			\end{tabular}
			
			\begin{tablenotes}
				\footnotesize
				\item[1]. FPS displays results on GTX 1050Ti device 
				\item[*]. Interactive double-branch 
				\item[+]. Interactive double-branch  + Weight multi-scale convolution 
			\end{tablenotes}
	\end{threeparttable}}
\end{table}

\textbf{WmConv:} WmConv introduces weight parameters on the basis of multi-scale convolution to capture feature information of multi-scale objects through multi-scale convolution kernels, reducing the influence of scale differences of remote sensing objects. At the same time, it is highlighed the differences between objects and backgrounds for suppressing complex background noise by assigning different weights to feature maps of sizes in multiple sensory field. It is used this approach to enhance the extracted object features. It is shown in Table \ref{tab4} as to the comparison results between WmConv + IDB and IDB on the UCAS-AOD object detection dataset under the same conditions. It is clear that using the WmConv module improves the mAP of the model by 0.88$\%$ compared with IDB, with the detection effect for small objects greatly improved from a single object ap.
\section{Conclusion }\label{sec5}
In this paper,we propose the Anchor Free rotating object detector BWP-Det. As per the characteristics of remote sensing objects and Anchor Free, a interactive double-branch up-sampling network is designed. At the same time, pixel level attention features are fused to guide the up-sampling process to highlight object features and weaken noise information. In addition, we, for the regression of boundary box parameters, designed a directional IoU loss function JIoU based on polar coordinate segmentation to jointly learn the length and width and angle parameters and avoid the periodic problem of angles. The ablation experiments showed the performance of each component of BWP-Det. It will be selected to equipment with stronger performance in future research. We will select higher baseline to improve the performance predicted by the model.
\bibliographystyle{elsarticle-num-names}
\bibliography{bibliography}
\end{document}